\PassOptionsToPackage{table,x11names}{xcolor}
\documentclass{article}

\PassOptionsToPackage{numbers, compress}{natbib}

    \usepackage[preprint]{neurips_2025}

\usepackage{tcolorbox}

\usepackage[utf8]{inputenc} 
\usepackage[T1]{fontenc}    
\usepackage{hyperref}       
\usepackage{url}            
\usepackage{booktabs}       
\usepackage{amsfonts}       
\usepackage{nicefrac}       
\usepackage{microtype}      
\usepackage{xcolor}       

\usepackage{hyperref}       
\hypersetup{
colorlinks= true,
urlcolor = blue,
linkcolor = red,
citecolor = blue,
}

\usepackage{xspace}
\usepackage{graphicx}
\usepackage{multicol}
\usepackage{multirow}
\usepackage{makecell}
\PassOptionsToPackage{numbers}{natbib}
\usepackage{wrapfig}
\usepackage{subcaption}
\usepackage{enumitem}
\usepackage{titletoc}
\usepackage{amsmath}
\usepackage{amssymb}
\usepackage{booktabs}    
\usepackage{longtable}   
\usepackage{array}       
\usepackage{pifont}

\usepackage{listings}
\newcommand{\keywords}[1]{%
  \vspace{1em}
  \noindent\textbf{Keywords:} #1
}

\setlist[itemize]{leftmargin=*}

\title{Knowledge-based Graphical Method for Safety Signal Detection in Clinical Trials}

\author{%
\textbf{Francois Vandenhende$^{1}$ \quad Anna Georgiou$^{1}$ \quad Michalis Georgiou$^{1}$}\\
\textbf{Theodoros Psaras$^{1}$ \quad Ellie Karekla$^{1}$ \quad Elena Hadjicosta$^{1}$}\\[6pt]
$^1$ClinBAY Limited, Limassol, Cyprus\\
Correspondence: \texttt{francois@clinbay.com}\\
\url{https://app.clinbay.com/safeterm}
}

\PassOptionsToPackage{numbers}{natbib}

\begin{document}

\maketitle

\begin{abstract}
We present a graphical, knowledge-based method for reviewing treatment-emergent adverse events (AEs) in clinical trials. The approach enhances MedDRA by adding a hidden medical knowledge layer (Safeterm) that captures semantic relationships between terms in a 2-D map. Using this layer, AE Preferred Terms can be regrouped automatically into similarity clusters, and their association to the trial disease may be quantified.
The Safeterm map is available online and connected to aggregated AE incidence tables from ClinicalTrials.gov. For signal detection, we compute treatment-specific disproportionality metrics using shrinkage incidence ratios. Cluster-level EBGM values are then derived through precision-weighted aggregation.
Two visual outputs support interpretation: a semantic map showing AE incidence and an expectedness-versus-disproportionality plot for rapid signal detection.
Applied to three legacy trials, the automated method clearly recovers all expected safety signals.
Overall, augmenting MedDRA with a medical knowledge layer improves clarity, efficiency, and accuracy in AE interpretation for clinical trials.
\end{abstract}


\section{Introduction}
The evaluation of treatment-emergent adverse events (AEs) is a central component of safety analysis in clinical trials. Traditional approaches rely heavily on tabular summaries of MedDRA-coded events, supplemented by selected graphical displays such as volcano plots or forest plots \cite{atypon2013_ae_volcano, phillips2020_visualisations_ae}. While these methods are widely adopted \cite{ichE2A, ichE9}, they often provide limited support for uncovering latent structure in AE data or for identifying nuanced safety patterns across related medical concepts \cite{nguyen2022_visual_harms}. As a result, safety reviewers may miss important semantic relationships between Preferred Terms (PTs), particularly when large and heterogeneous AE profiles are involved.

MedDRA itself provides a rich and hierarchical vocabulary for AE coding, but it does not explicitly encode clinical knowledge beyond its predefined parent--child taxonomy \cite{ich1999_meddra_structure, mozzicato2009_meddra_overview}. Common aggregation strategies, such as using Standardised MedDRA Queries (SMQs) \cite{mozzicato2007_smq_signal, chang2017_smq_usage}, FDA Office of Custom Medical Queries \cite{fdaOCMQ_article2025}, SOC-level or hybrid \cite{dupuch2012_hybrid_grouping} summaries, offer broad grouping, but do not fully reflect latent semantic or mechanistic relationships specific to a drug, indication, or clinical trial. Consequently, analyses that treat MedDRA strictly as a hierarchical dictionary or using predefined groupings of terms may fail to capture deeper medical connections that could enhance AE interpretation and signal identification.

In response to these limitations, the intersection of machine learning, natural language processing, and Bayesian inference has introduced powerful methods to enrich AE evaluation. Embedding-based models and semantic similarity measures are gaining traction in pharmacovigilance for their ability to represent conceptual relationships beyond rigid hierarchies \cite{duMouchelEBGM}, combine terms in relevant groups \cite{painter2025_ssm_clustering} or improve disproportionality assessment in PV using information sharing \cite{haguinet2025_icssm_bdb}.

These emerging approaches are promising but have seen limited application in clinical trial settings, where AE data are often sparse and more structured than post‑marketing spontaneous reports. Disproportionality methods such as Empirical Bayes Geometric Mean (EBGM) or Bayesian hierarchical models remain standard in signal detection \cite{duMouchelEBGM, noren2013}, yet typically ignore semantic relationships among PTs. Similarly, statistical tools for safety review still largely rely on hierarchical groupings or co‑occurrence statistics \cite{wang2022safetyviz}.

In this work, we introduce a knowledge‑based graphical framework that augments MedDRA with an additional, hidden medical knowledge layer. This layer, encoded in the Safeterm system \cite{safetermapp}, provides a set of semantic descriptors and inter-term relationships that reflect clinical reasoning and knowledge not captured in MedDRA’s hierarchy. 
Our method combines high-dimensional semantic embeddings of MedDRA PTs, with a set of downstream applications tailored for clinical trial safety review: semantic clustering, Bayesian shrinkage disproportionality metrics at both the PT and cluster levels, and indication-specific expectedness scores by comparing PT embeddings to the semantic representation of the patient population. 

To ease trial safety review, we present two intuitive visualizations: a two-dimensional semantic map of AEs and an expectedness-versus-disproportionality plot.

The addition of a hidden knowledge layer to MedDRA offers a context-aware, interpretable, and scalable enhancement to conventional safety signal detection. It provides an effective way to identify clinically meaningful safety patterns and reduce the manual burden of AE review in trials.

\section{Methods}

\subsection{Hidden Medical Knowledge Layer}
The hidden medical knowledge layer, implemented in the \textit{Safeterm} system, encodes MedDRA terminology into high-dimensional numerical vectors, or embeddings. These embeddings capture semantic, clinical, and contextual relationships among Preferred Terms (PTs). PTs with similar meanings or shared mechanistic characteristics occupy nearby positions in this space, enabling multivariate analyses to explore term dependencies and latent structures that are not explicitly encoded in MedDRA’s hierarchy. 

While this high-dimensional representation is useful for computational analyses, it is not directly interpretable. Therefore, a set of decoding tools was developed to project signals back onto the MedDRA domain, supporting human interpretability.

\subsection{Decoding Toolbox for PT Understanding}
The decoding toolbox currently consists of three key components: PT clustering, two-dimensional interpretability projection, and expectedness scoring. These tools are applied to the hidden layer to facilitate understanding, visualization, and prioritization of PTs. Additional tools will be added in future releases of the Safeterm app.

\subsubsection{Semantic Clustering of PTs}

The embeddings of observed PTs in each trial were first projected into a reduced–dimensional space using Principal Component Analysis (PCA) \cite{jolliffe2016pca}.  
This step preserves the most informative variance in the embedding space while improving cluster separability, computational efficiency, and identifiability of downstream similarity structures.
Next, we applied a clustering algorithm to the PCA‑reduced embeddings in order to identify groups of semantically related PTs and extract them from the random noise. 

Cluster naming was performed using an AI-based decoding method that assigns interpretable labels to clusters based on their content centroid in the hidden medical knowledge space. Terms that did not group into any cluster were flagged in our reports for transparency. 

\subsubsection{Safeterm Map: Two-Dimensional Interpretability Projection}
To enable interactive exploration, the high-dimensional embeddings were projected into a two-dimensional space for visualization. Each PT is represented as a point, and proximity between points reflects semantic similarity. Incidence data from the clinical trial were overlaid on the map by treatment arm, highlighting clustering patterns and enabling the identification of treatment-specific mapping differences. The Safeterm map facilitates intuitive exploration of AE relationships and provides a context-aware view of trial safety data.

\subsubsection{Indication-Specific Expectedness Score}
Patient population descriptors were extracted from the clinical protocol and mapped into the hidden space using the Safeterm encoder. Cosine similarity scores between the patient population descriptors and PT embeddings were computed, providing an expectedness score for each PT. Cosine similarity is widely used in natural language processing and biomedical embeddings \cite{sanchez2011semantic, painter2025_ssm_clustering}. Higher scores indicate that a PT is more likely to be observed given the trial population. This measure allows safety reviewers to contextualize AE occurrences relative to the expected background profile of the trial population.

\subsection{Statistical methods}

	The method uses aggregated incidence tables of subjects with adverse events, by Preferred Term (PT) and treatment group, consistent with standard clinical trial AE reporting practice. 
	To identify potential safety signals beyond simple incidence counts, we computed the \textbf{Empirical Bayes Geometric Mean (EBGM)} statistics, which provide a robust disproportionality measure adjusted for sparse event counts \cite{openEBGM2017}.
	
	For each PT, let $n_i$ denote the observed count of subjects with events and $N_i$ the total number of subjects at risk under treatment $i$ ($i=1,...,T$).
	The \textit{expected count of events} $E_i$ for treatment $i$ is calculated as the product of the treatment sample size by the overall incidence of the PT across all treatment arms:
	
	\begin{equation}
	\nonumber
	E_i = N_i \cdot \frac{\sum_{i} n_i}{\sum_{\text{i}} N_i}.
	\end{equation}
	
	\noindent
	This choice of denominator differs from classical pharmacovigilance methods, which often use an external null or historical incidence \cite{noren2013}. Using the trial-specific overall incidence allows for an \textit{internally consistent background} for relative comparisons between treatment arms, simplifies automated processing, and introduces only minor loss for exploratory signal detection.
	
	The EBGM was then computed with prior stabilization as:
	
	\begin{equation}
		\nonumber
	EBGM_i = \frac{n_i + \alpha}{E_i + \beta},
	\end{equation}
	
	\noindent
	where $\alpha$ and $\beta$ are prior parameters that reduce variance for low-count PTs. The posterior distribution of EBGM can then be obtained using conjugated gamma priors \cite{openEBGM2017}.
	
	\subsubsection{Cluster-level EBGM}

To integrate semantic information from the hidden Safeterm space, EBGM values were aggregated at the cluster level.  
Because patient-level information was not available, a meta-analytic weighting strategy was applied.

For each Preferred Term (PT) \( j \) within a cluster, the sampling variance was approximated as:

\begin{equation}
\text{Var}_{\text{EBGM},j} = \frac{\text{EBGM}_j^2}{n_j + \alpha},
\end{equation}

\noindent
where \( n_j \) is the number of subjects reporting the PT and \( \alpha \) is the prior parameter used in the EBGM shrinkage model.  
The precision was then defined as:

\begin{equation}
w_j = \frac{1}{\text{Var}_{\text{EBGM},j}}.
\end{equation}

The cluster-level EBGM for each treatment group was computed as a precision-weighted average:

\begin{equation}
\text{EBGM}_{\text{cluster}} = 
\frac{\sum_{j} w_j \cdot \text{EBGM}_j}{\sum_{j} w_j}.
\end{equation}

This approach stabilizes estimates for rare events while preserving semantic coherence among PTs within a cluster \cite{painter2025_ssm_clustering}.  
Although future extensions could quantify uncertainty at the cluster level (e.g., via Bayesian hierarchical modeling), this study focuses on point estimates only, reflecting its exploratory safety-review objective.  
Overall, the method provides a practical and interpretable summary measure for cluster-level disproportionality.

\subsubsection{Expectedness versus Disproportionality Plot}
	
	To contextualize signals relative to trial-specific expectations, we developed the Expectedness versus Disproportionality (EVD) plot. In this plot:
	
	\begin{itemize}
		\item The x-axis represents the \textit{expectedness score} of each PT, determining how the PT relates to the studied indication, or patient descriptors.
		\item The y-axis represents the \textit{EBGM value}, reflecting how the AE incidence compares in a treatment arm versus in the entire study population.
	\end{itemize}
	
	PTs are grouped according to Safeterm clusters. This visualization provides a synthetic, interpretable overview of major safety patterns, highlighting PTs and clusters that are both unexpected and disproportionately observed in a treatment arm.
	\subsection{Performance Evaluation}

	The performance of the proposed method was explored using three clinical trials with publicly available AE data from ClinicalTrials.gov. For each study, the incidence of subjects with AEs was computed as the sum of counts for serious and other (non-serious) events. Only AEs coded as valid MedDRA PTs were included.

	The selected trials were: 
	\begin{enumerate}
		\item \textbf{NCT05096221} – Phase 3, randomized, double-blind, placebo-controlled study evaluating SRP-9001 in Duchenne Muscular Dystrophy (DMD) \cite{NCT05096221}.
		\item \textbf{NCT02348593} – Twelve-week, double-blind, placebo-controlled, randomized study of JZP-110 in narcolepsy \cite{NCT02348593}.
		\item \textbf{NCT05008224} – Phase 2 sequential oncology study of pembrolizumab and chemotherapy in classical Hodgkin lymphoma (KEYNOTE-C11) \cite{NCT05008224}.
	\end{enumerate}

	These trials were chosen to illustrate the method across different therapeutic areas, patient populations, and AE profiles. NCT05096221 involves a gene therapy in DMD with known transient liver AEs; NCT02348593 is a dose-dependent stimulant study in narcolepsy; and NCT05008224 is a sequential oncology trial with non-randomized chemotherapy groups, decided based on PET scan outcome. It illustrates treatment-safety confounding by disease severity.

	\section{Results}	
	
	\subsection{Duchenne Muscular Dystrophy Trial}

This phase III trial reported a total of 72 distinct Preferred Terms (PTs), of which 44 were clustered into 7 Safeterm groups. Three main groups are detailed in Table~\ref{tab:dmd_results}, combining PTs from multiple SOCs.  

A notable cluster labeled \emph{Liver damage} was identified, encompassing PTs such as \emph{Elevated liver function tests}, \emph{Hepatic enzyme increased}, and \emph{Liver injury}. This cluster captured the spectrum of liver-related adverse events (AEs) observed in the trial, facilitating a more comprehensive safety analysis.

\begin{figure}
		\centering
		\includegraphics[width=1.0\textwidth]{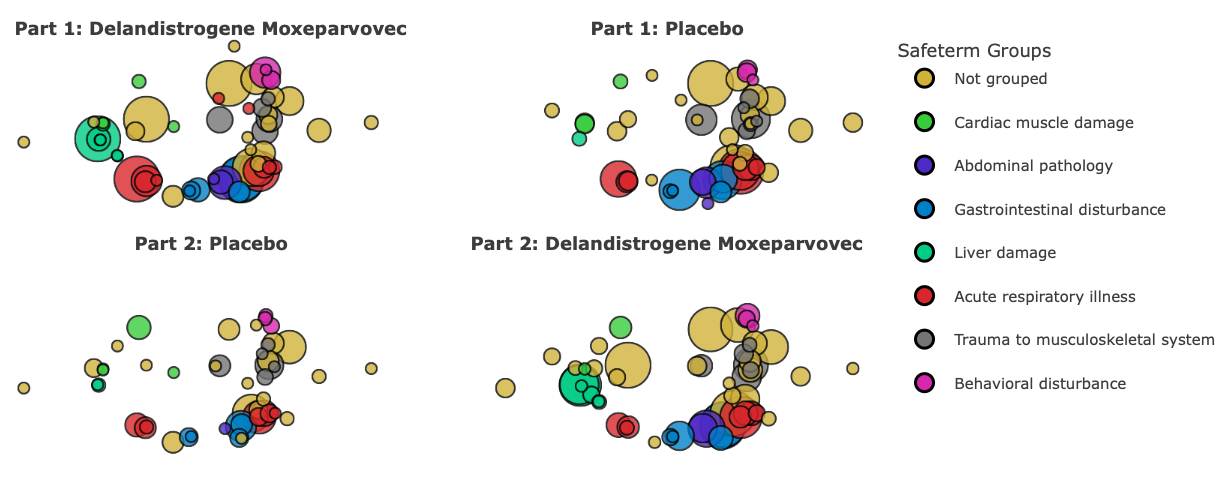}
		\caption{Safeterm map of incidence of subjects with AEs by PT, treatment and cluster for the DMD trial.}
		\label{fig:fig_01}
	\end{figure}

Figure~\ref{fig:fig_01} presents the Safeterm 2-D map for the reported PTs. Each dot represents a PT, with its size proportional to the proportion of subjects affected. PTs are colored by Safeterm cluster, with ungrouped terms displayed in brown. Clusters correspond to PTs that occupy similar coordinates in the projected space. For example, the \emph{Liver damage} cluster is located in the central-left quadrant of the map. The map shows more and larger dots for liver-related AEs in the active treatment groups compared to placebo. This visual pattern suggests a treatment-related increase in liver AEs.

\begin{figure}
		\centering
		\includegraphics[width=1.0\textwidth]{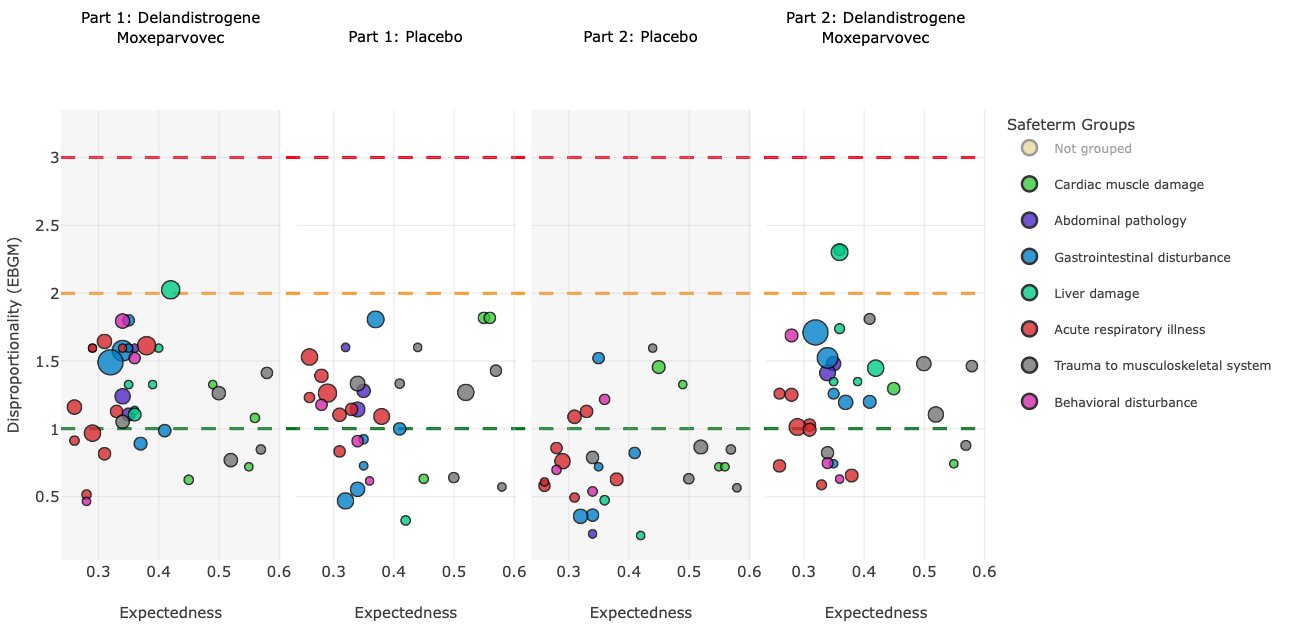}
		\caption{EVD plot of incidence of subjects with AEs by treatment for clustered PTs (DMD trial).}
		\label{fig:fig_02}
	\end{figure}

Figure~\ref{fig:fig_02} shows the Expectedness versus Disproportionality (EVD) plot for the clustered PTs. The x-axis represents the expectedness score for Duchenne muscular dystrophy, while the y-axis shows the Empirical Bayes Geometric Mean (EBGM) reflecting disproportionality relative to the overall trial population. PTs associated with musculoskeletal trauma have the highest expectedness, whereas terms from acute respiratory illness (e.g., \emph{Nasopharyngitis}, \emph{Epistaxis}) have the lowest. Each dot is a PT, with size proportional to incidence.

Focusing on the \emph{Liver damage} cluster, all terms show a consistent EBGM greater than 1 for active treatment versus placebo. PTs such as \emph{GGT increased} and \emph{Blood bilirubin increased} demonstrate the highest disproportionality, with EBGM point estimates above 2, indicating a strong association with treatment. This observation is corroborated by the cluster-level EBGM analysis (see Table~\ref{tab:ebgm_results}), which provides a comparative treatment overview across all PTs forming relevant clusters.

	\subsection{Dose-Response Narcolepsy Trial}

The second study illustrates the utility of our method in detecting dose-dependent pharmacological effects. 

A total of 18 distinct Preferred Terms (PTs) were reported in this trial, of which 15 were clustered into 2 categories: \emph{Stress response} and \emph{Respiratory infection}.  

The \emph{Stress response} cluster included 12 PTs such as \emph{Increased heart rate}, \emph{Anxiety}, \emph{Nervousness}, and \emph{Insomnia}, reflecting heightened sympathetic activity consistent with the pharmacological action of JZP-110 in narcolepsy. 

\begin{figure}
		\centering
		\includegraphics[width=1.0\textwidth]{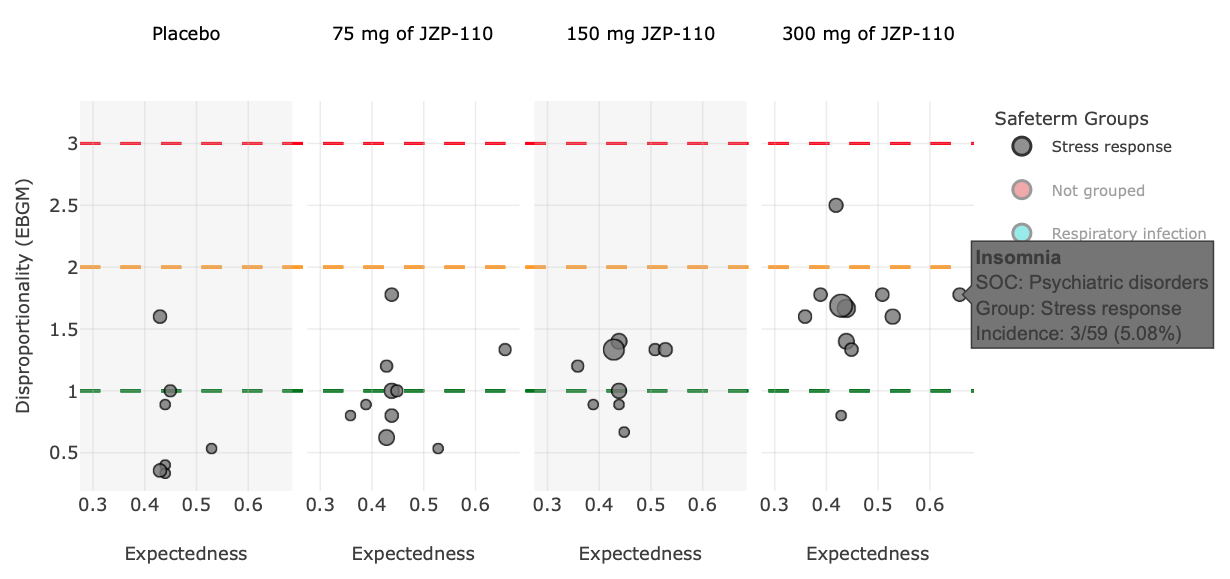}
		\caption{EVD plot of incidence of subjects with AEs by PT, treatment and cluster for the narcolepsy trial.}
		\label{fig:fig_03}
	\end{figure}

This cluster shows the strongest connection to the disease and treatment mechanism (see Figure~\ref{fig:fig_03}). A clear monotonic dose-response relationship is observed, with the incidence of stress-related AEs increasing with higher JZP-110 doses (see Table~\ref{tab:results2}). Most stress-related events were non-serious, with the exception of a single episode of \emph{Anxiety} at the 150 mg dose.  

These results demonstrate that our augmented MedDRA analyses can identify PT patterns beyond standard safety, aligned with known pharmacological effects of drugs, supporting both safety and mechanistic interpretations in clinical trials.

	\subsection{Hodgkin's Lymphoma Trial}

\begin{figure}
		\centering
		\includegraphics[width=1.0\textwidth]{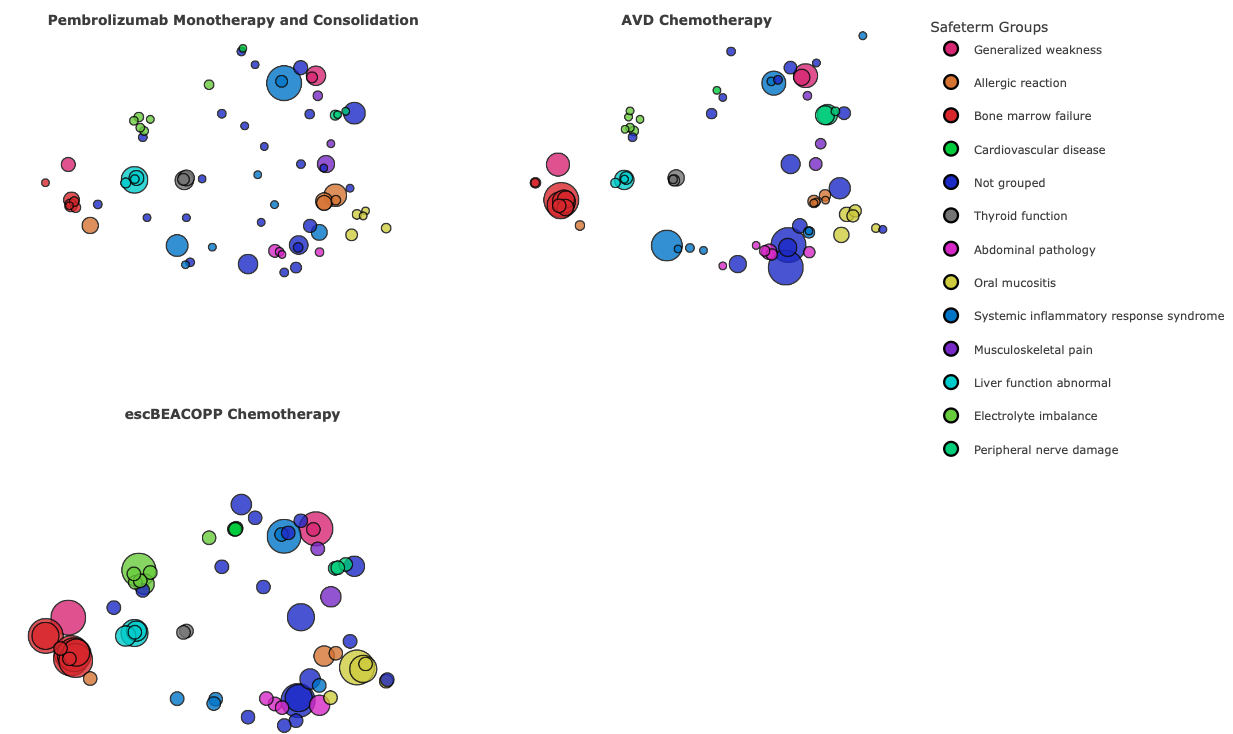}
		\caption{Safeterm map of incidence of subjects with AEs by PT, treatment and cluster for the lymphoma trial}
		\label{fig:fig_04}
	\end{figure}

	In this sequential oncology trial, a total of 104 distinct Preferred Terms (PTs) were reported, which clustered into 12 identifiable semantic groups (see Figure~\ref{fig:fig_04}).  
	
\begin{figure}
		\centering
		\includegraphics[width=1.0\textwidth]{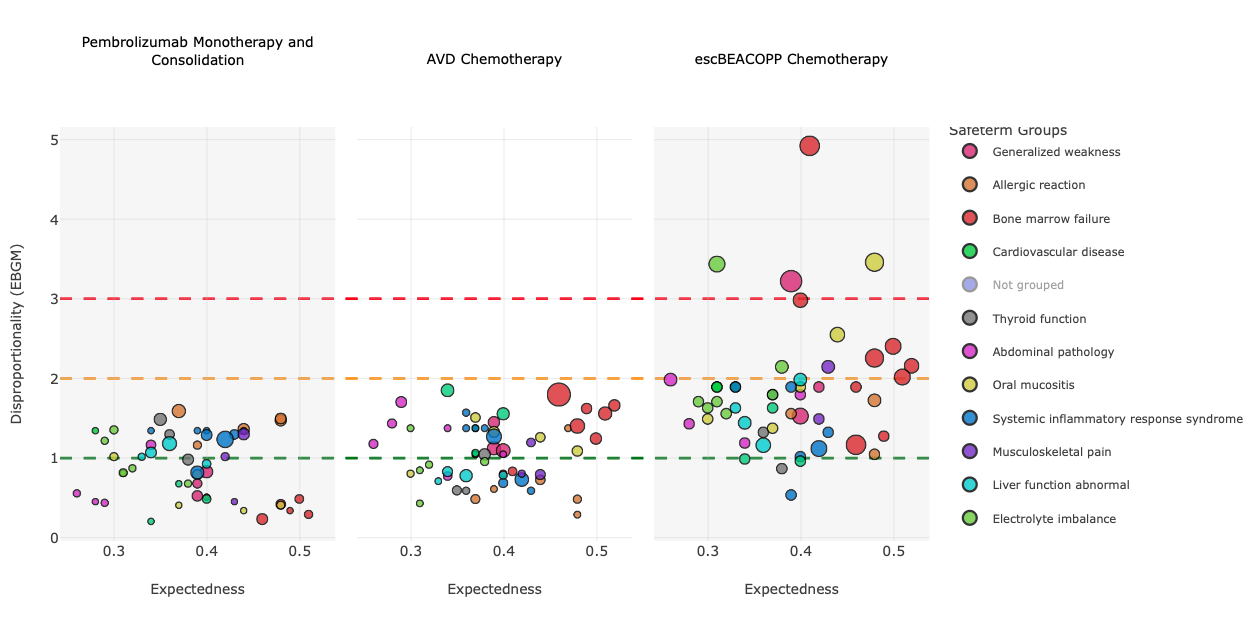}
		\caption{EVD plot of incidence of subjects with AEs by PT, treatment and cluster for the lymphoma trial}
		\label{fig:fig_05}
	\end{figure}

	As shown in Figure~\ref{fig:fig_05}, the \emph{Bone marrow failure} cluster was most closely associated with the disease and its treatment. Patients receiving escBEACOPP, selected based on unfavorable PET scan outcomes, exhibited a higher incidence of bone marrow–related adverse events compared to those receiving AVD chemotherapy or Pembrolizumab monotherapy.  
	
	Despite the trial’s non-randomized design and unbalanced treatment group sizes, our method provided a rapid and informative overview of safety signals contextualized to the disease. The tri-variate association among patient characteristics, treatment, and adverse events enabled clearer interpretation of safety signals in this complex study setting.
	
	\section{Discussion}

	This study demonstrates the potential of augmenting MedDRA terminology with a hidden medical knowledge layer to enhance safety and pharmacology signal detection in clinical trials. We applied the approach to three distinct trials, each illustrating different challenges in safety review: a gene therapy trial in Duchenne Muscular Dystrophy (DMD) with known liver-related adverse events (AEs); a dose-response study in narcolepsy where a mechanistic effect was detected; and a sequential oncology trial where treatment and disease population are confounded.
	
	In all cases, the combination of automated term grouping, AE-disease expectedness evaluation, and two graphical displays (Safeterm Map and EVD plot) proved highly effective in identifying relevant safety signals. The method provided a rapid and intuitive overview of AE patterns, demonstrating clear advantages over traditional inspection of AE tables.
	
	While these results are promising, further work is needed to systematically quantify performance improvements across a broader range of clinical trials. Future research could incorporate additional AE attributes, such as severity, duration, and onset time. Although the current method is exploratory, further development is needed to enable causal inference, either at the study level or across trials.  
	
	Individual patient-level data could also be analyzed, including medical history, concomitant medications, and AEs. These could be mapped onto the medical knowledge space to better understand their relationships and reveal drug-disease-event interactions.

	\section{Conclusion}

Augmenting MedDRA with a hidden medical knowledge layer enhances the analysis of adverse events in clinical trials. Semantic clustering, 2-D interpretability maps, and Expectedness versus Disproportionality plots allow reviewers to identify clinically meaningful AE patterns efficiently and in context.  

Applied to diverse trials, including gene therapy, dose-response, and oncology studies, the method facilitated rapid detection of treatment-specific safety signals and highlighted relevant pharmacological effects. The Safeterm online app, connected to \texttt{clinicaltrials.gov}, provides an accessible, interactive platform for exploring trial AE data, supporting more informed and streamlined safety review.

\section*{Statements and Declarations}

\textbf{Funding:} No funding was provided for this study.\\[4pt]

\textbf{Conflicts of interest:} 
The authors are affiliated with ClinBAY Ltd; they declare no financial or commercial conflict beyond this.\\[4pt]

\textbf{Author contributions:} 
All authors contributed equally to concept, methodology, data analysis, and writing.\\[4pt]

\textbf{Ethics approval:} Not applicable; retrospective analysis of existing data.\\[4pt]

\textbf{Data availability:}\\
\url{https://clinicaltrials.gov/study/NCT05096221}\\
\url{https://clinicaltrials.gov/study/NCT02348593}\\
\url{https://clinicaltrials.gov/study/NCT05008224}\\

\bibliographystyle{unsrtnat}
\bibliography{main}

@article{atypon2013_ae_volcano,
  author    = {Zink, Richard C. and Wolfinger, Russell D. and Mann, Geoffrey},
  title     = {Summarizing the incidence of adverse events using volcano plots and time intervals},
  journal   = {Clinical Trials},
  year      = {2013},
  volume    = {10},
  number    = {3},
  pages     = {398--406},
  doi       = {10.1177/1740774513485311},
  pmid      = {23690094}
}

@article{phillips2020_visualisations_ae,
  author    = {Cornelius, Victoria and Cro, Suzie and Phillips, Rachel},
  title     = {Advantages of visualisations to evaluate and communicate adverse event information in randomised controlled trials},
  journal   = {Trials},
  year      = {2020},
  volume    = {21},
  number    = {1},
  pages     = {1028},
  doi       = {10.1186/s13063-020-04903-0},
  pmid      = {33353566},
  pmcid     = {PMC7754702}
}

@techreport{ichE2A,
  title       = {{ICH E2A: Clinical Safety Data Management: Definitions and Standards for Expedited Reporting}},
  author      = {{International Council for Harmonisation of Technical Requirements for Pharmaceuticals for Human Use (ICH)}},
  institution = {ICH},
  year        = {1994},
  type        = {Guideline},
  url         = {https://www.ich.org/page/efficacy-guidelines},
  note        = {Accessed 2025-11-19}
}

@techreport{ichE9,
  title       = {{ICH E9: Statistical Principles for Clinical Trials}},
  author      = {{International Council for Harmonisation of Technical Requirements for Pharmaceuticals for Human Use (ICH)}},
  institution = {ICH},
  year        = {1998},
  type        = {Guideline},
  url         = {https://www.ich.org/page/efficacy-guidelines},
  note        = {Accessed 2025-11-19}
}

@article{nguyen2022_visual_harms,
  author    = {Qureshi, Riaz and Chen, Xiwei and Goerg, Carsten and Mayo-Wilson, Evan and Dickinson, Stephanie and Golzarri-Arroyo, Lilian and Hong, Hwanhee and Phillips, Rachel and Cornelius, Victoria and McAdams DeMarco, Mara and Guallar, Eliseo and Li, Tianjing},
  title     = {Comparing the Value of Data Visualization Methods for Communicating Harms in Clinical Trials},
  journal   = {Epidemiologic Reviews},
  year      = {2022},
  volume    = {44},
  number    = {1},
  pages     = {55--66},
  doi       = {10.1093/epirev/mxac005},
  pmid      = {36065832},
  pmcid     = {PMC9780120}
}

@article{ich1999_meddra_structure,
  author    = {Brown, E. G. and Wood, L. and Wood, S.},
  title     = {The Medical Dictionary for Regulatory Activities (MedDRA)},
  journal   = {Drug Safety},
  year      = {1999},
  volume    = {20},
  number    = {2},
  pages     = {109--117},
  doi       = {10.2165/00002018-199920020-00002},
  pmid      = {10082069}
}

@article{mozzicato2009_meddra_overview,
  author       = {Mozzicato, Patricia},
  title        = {{MedDRA: An Overview of the Medical Dictionary for Regulatory Activities}},
  journal      = {Pharmaceutical Medicine},
  year         = {2009},
  volume       = {23},
  number       = {2},
  pages        = {65--75},
  doi          = {10.1007/BF03256752},
  url          = {https://doi.org/10.1007/BF03256752}
}

@article{mozzicato2007_smq_signal,
  author       = {Mozzicato, Patricia},
  title        = {Standardised MedDRA Queries: Their Role in Signal Detection},
  journal      = {Drug Safety},
  year         = {2007},
  volume       = {30},
  number       = {7},
  pages        = {617--619},
  doi          = {10.2165/00002018-200730070-00009},
  url          = {https://www.ovid.com/journals/drusa/fulltext/00002018-200730070-00009}
}

@article{chang2017_smq_usage,
  author       = {Chang, Lin Chau and Mahmood, Riaz and Qureshi, Samina and Breder, Christopher D.},
  title        = {Patterns of Use and Impact of Standardised MedDRA Query Analyses on the Safety Evaluation and Review of New Drug and Biologics License Applications},
  journal      = {PLOS ONE},
  year         = {2017},
  volume       = {12},
  number       = {6},
  pages        = {e0178104},
  doi          = {10.1371/journal.pone.0178104},
  url          = {https://doi.org/10.1371/journal.pone.0178104}
}

@article{fdaOCMQ_article2025,
  author    = {Proestel, Scott and Popat, Vaishali and Unger, Ellis F. and Jeng, Linda J. B.},
  title     = {The Development and Use of Office of New Drugs Custom Medical Queries for Safety Analyses of Clinical Trial Data},
  journal   = {Drug Safety},
  year      = {2025},
  volume    = {48},
  number    = {9},
  pages     = {1331--1337},
  doi       = {10.1007/s40264-025-01582-1},
  url       = {https://doi.org/10.1007/s40264-025-01582-1}
}

@article{dupuch2012_hybrid_grouping,
author    = {Dupuch, Marie and Dupuch, La{\"e}titia and Perinet, Amandine and Hamon, Thierry and Grabar, Natalia},
  title     = {Grouping the pharmacovigilance terms with a hybrid approach},
  journal   = {Studies in Health Technology and Informatics},
  year      = {2012},
  volume    = {180},
  pages     = {235--239},
  pmid      = {22874187},
  note      = {Combines semantic similarity and terminology structuring; references SMQs explicitly},
  url       = {https://pubmed.ncbi.nlm.nih.gov/22874187}
}

@article{duMouchelEBGM,
  author       = {DuMouchel, William},
  title        = {Bayesian Data Mining in Large Frequency Tables, with an Application to the FDA Spontaneous Reporting System},
  journal      = {The American Statistician},
  year         = {1999},
  volume       = {53},
  number       = {3},
  pages        = {177--190},
  doi          = {https://doi.org/10.1080/00031305.1999.10474456}
}

@article{painter2025_ssm_clustering,
  author        = {Painter, Jeffery L. and Haguinet, Fran\c{c}ois and Powell, Gregory E. and Bate, Andrew},
  title         = {Ontology-based Semantic Similarity Measures for Clustering Medical Concepts in Drug Safety},
  year          = {2025},
  archivePrefix = {arXiv},
  eprint        = {2503.20737},
  primaryClass  = {cs.AI},
  note          = {Preprint from GlaxoSmithKline}
}

@article{haguinet2025_icssm_bdb,
  author        = {Haguinet, Fran\c{c}ois and Painter, Jeffery L. and Powell, Gregory E. and Callegaro, Andrea and Bate, Andrew},
  title         = {Bayesian Dynamic Borrowing Considering Semantic Similarity Between Outcomes for Disproportionality Analysis},
  year          = {2025},
  archivePrefix = {arXiv},
  eprint        = {2504.12052},
  primaryClass  = {stat.ME},
  note          = {Preprint from GlaxoSmithKline (IC SSM approach)}
}

@article{noren2013,
  author    = {Nor{\'e}n, G. Niklas and Sundberg, Rolf and Bate, Andrew and Edwards, I. Ralph},
  title     = {A statistical methodology for drug--drug interaction surveillance},
  journal   = {Statistics in Medicine},
  year      = {2008},
  volume    = {27},
  number    = {16},
  pages     = {3057--3070},
  doi       = {10.1002/sim.3247},
  pmid      = {18344185},
  note      = {Erratum published in Statistics in Medicine, 2008; 27(29):6271--6272}
}

@article{wang2022safetyviz,

  author    = {de Abreu Nunes, Laetitia and Hooper, Richard and McGettigan, Patricia and Phillips, Rachel},
  title     = {Statistical methods leveraging the hierarchical structure of adverse events for signal detection in clinical trials: a scoping review of the methodological literature},
  journal   = {BMC Medical Research Methodology},
  year      = {2024},
  volume    = {24},
  number    = {1},
  pages     = {253},
  doi       = {10.1186/s12874-024-02369-1},
  pmid      = {39468481},
  pmcid     = {PMC11514772}
}

@misc{safetermapp,
  title        = {{Safeterm: Interactive Safety Signal Review Tool}},
  howpublished = {\url{https://app.clinbay.com/safeterm}},
  note         = {Accessed 2025-11-20},
  year         = {2025},
  author       = {{ClinBay Ltd.}}
}

@book{jolliffe2016pca,
  title        = {Principal Component Analysis},
  author       = {Jolliffe, I. T. and Cadima, J.},
  journal      = {Philosophical Transactions of the Royal Society A: Mathematical, Physical and Engineering Sciences},
  year         = {2016},
  volume       = {374},
  pages        = {20150202},
  doi          = {10.1098/rsta.2015.0202},
  note         = {Comprehensive review of PCA, including applications for dimensionality reduction of high-dimensional embeddings.}
}

@article{sanchez2011semantic,
  title={Semantic similarity estimation in the biomedical domain: an ontology-based information-theoretic perspective},
  author={Sanchez, Daniel and Batet, Montse},
  journal={J Biomed Inform},
  volume={44},
  number={5},
  pages={749--759},
  year={2011},
  doi={10.1016/j.jbi.2011.03.013},
  url={https://www.ncbi.nlm.nih.gov/pubmed/21463704}
}

@article{openEBGM2017,
  author = {Canida, Travis and Ihrie, John},
  title = {openEBGM: An R Implementation of the Gamma-Poisson Shrinker Data Mining Model},
  journal = {The R Journal},
  year = {2017},
  note = {https://doi.org/10.32614/RJ-2017-063},
  doi = {10.32614/RJ-2017-063},
  volume = {9},
  issue = {2},
  issn = {2073-4859},
  pages = {499-519}
}

@article{NCT05096221,
  author       = {Mendell, Jerry R. and Muntoni, Francesco and McDonald, Craig M. and Mercuri, Eugenio M. and Ciafaloni, Emma and Komaki, Hirofumi and Leon‑Astudillo, Carmen and Nascimento, Andrés and Proud, Crystal and Schara‑Schmidt, Ulrike and Veerapandiyan, Aravindhan and Zaidman, Craig M. and Guridi, Maitea and Murphy, Alexander P. and Reid, Carol and Wandel, Christoph and Asher, Damon R. and Darton, Eddie and Mason, Stefanie and Potter, Rachael A. and Singh, Teji and Zhang, Wenfei and Fontoura, Paulo and Elkins, Jacob S. and Rodino‑Klapac, Louise R.},
  title = {{AAV} gene therapy for {D}uchenne muscular dystrophy: the {EMBARK} phase 3 randomized trial},
  journal      = {Nature Medicine},
  year         = {2025},
  volume       = {31},
  number       = {1},
  pages        = {332--341},
  doi          = {10.1038/s41591-024-03304-z},
  note         = {ClinicalTrials.gov Identifier: NCT05096221}
}

@article{NCT02348593,
  author    = {Rosenberg, Russell and Thorpy, Michael J. and Dauvilliers, Yves and Schweitzer, Paula K. and Zammit, Gary and Gotfried, Mark and Bujanover, Shay and Scheckner, Brian and Malhotra, Atul},
  title     = {Incidence and duration of common early-onset adverse events in randomized controlled trials of solriamfetol for treatment of excessive daytime sleepiness in obstructive sleep apnea and narcolepsy},
  journal   = {Journal of Clinical Sleep Medicine},
  year      = {2022},
  volume    = {18},
  number    = {1},
  pages     = {235--244},
  doi       = {10.5664/jcsm.9550},
  pmid      = {34283019},
  pmcid     = {PMC8807921}
}

@article{NCT05008224,
author = {Advani, Ranjana and Avigdor, Abraham and Balari, Anna and Lavie, David and Hohaus, Stefan and Zaucha, Jan and Hua, Vu and Zilioli, Vittorio and Gazitua, Raimundo and Ozcan, Muhit and Odeleye-Ajakaye, Amos and Reddy, Nishitha and Marinello, Patricia and Winter, Jane},
year = {2023},
month = {11},
pages = {3068-3068},
title = {Efficacy and Safety of Pembrolizumab and Chemotherapy in Newly-Diagnosed, Early Unfavorable or Advanced Classic Hodgkin Lymphoma: The Phase 2 Keynote-C11 Study},
volume = {142},
journal = {Blood},
doi = {10.1182/blood-2023-181171}
}

\clearpage
\begin{table}[ht]
	\centering
	\caption{List of MedDRA PTs in Selected Safeterm Groupings for Duchene Muscular Dystrophy Trial}
	\label{tab:dmd_results}
	\renewcommand{\arraystretch}{1.2}
	\begin{tabular}{lrr}
	\toprule
	\textbf{SafeTerm Group} & \textbf{SOC} & \textbf{Term} \\ 
	\midrule
	
	Abdominal pathology & Infections and infestations & Appendicitis \\ 
	 & & Anal abscess \\ 
	 & Gastrointestinal disorders & Abdominal pain upper \\ 
	 & & Abdominal pain \\

	Gastrointestinal disturbance & Infections and infestations & Rotavirus infection \\ 
	 &  & Gastroenteritis viral \\ 
	 &  & Gastroenteritis \\ 
	 & Gastrointestinal disorders & Vomiting \\ 
	 &  & Nausea \\ 
	 &  & Diarrhoea \\ 
	 &  & Constipation \\ 
	
	Liver damage & Investigations & Transaminases increased \\ 
	 &  & Hepatic enzyme increased \\ 
	&  & Glutamate dehydrogenase increased \\ 
	 &  & Gamma-glutamyltransferase increased \\ 
	 &  & Blood bilirubin increased \\ 
	 &  & Liver injury \\ 
	 &  & Hepatotoxicity \\ 
	\bottomrule
	
	\end{tabular}
	\end{table}
	
\clearpage
\begin{table}[h!]
	\centering
	\caption{EBGM at the PT Group Level for Duchene Muscular Dystrophy Trial}
	\label{tab:ebgm_results}
	\begin{tabular}{lrrrrr}
		\toprule
		\textbf{Safeterm PT Group} & 
		\makecell{\textbf{Part 1:} \\ \textbf{Delandistrogene} \\ \textbf{Moxeparvovec}} & 
		\makecell{\textbf{Part 1:} \\ \textbf{Placebo}} & 
		\makecell{\textbf{Part 2:} \\ \textbf{Delandistrogene} \\ \textbf{Moxeparvovec}} & 
		\makecell{\textbf{Part 2:} \\ \textbf{Placebo}} \\
		\midrule
		Liver damage & 1.30 & 0.29 & 1.59 & 0.34 \\
		Gastrointestinal disturbance & 1.15 & 0.60 & 1.41 & 0.33 \\
		Abdominal pathology & 1.16 & 1.18 & 1.29 & 0.21 \\
		Acute respiratory illness & 1.03 & 1.15 & 0.85 & 0.71 \\
		Behavioral disturbance & 0.91 & 0.87 & 0.86 & 0.68 \\
		Trauma to musculoskeletal system & 0.89 & 0.99 & 1.07 & 0.73 \\
		Cardiac muscle damage & 0.78 & 0.92 & 0.69 & 0.98 \\
		\bottomrule
	\end{tabular}
\end{table}

\clearpage

\begin{table}[ht]
	\centering
	\caption{Incidence of Patients with AEs by PT and Treatment For Stress Response Cluster in Narcolepsy Trial}
	\label{tab:results2}
	\begin{tabular}{lllrr}
		\toprule
		\textbf{Term} & \textbf{150 mg JZP-110} & \textbf{300 mg JZP-110} & \textbf{75 mg JZP-110} & \textbf{Placebo} \\ 
		\midrule	
	
	 Insomnia & 0/59 (0.0\%) & 3/59 (5.08\%) & 2/59 (3.39\%) & 0/59 (0.0\%) \\ 
	  Anxiety & 4/59 (6.78\%) & 5/59 (8.47\%) & 1/59 (1.69\%) & 1/59 (1.69\%) \\ 
	   Headache & 14/59 (23.73\%) & 18/59 (30.51\%) & 6/59 (10.17\%) & 3/59 (5.08\%) \\ 
	 Dizziness & 1/59 (1.69\%) & 3/59 (5.08\%) & 2/59 (3.39\%) & 2/59 (3.39\%) \\ 
	 Decreased appetite & 5/59 (8.47\%) & 9/59 (15.25\%) & 5/59 (8.47\%) & 1/59 (1.69\%) \\ 
	 Weight increased & 0/59 (0.0\%) & 1/59 (1.69\%) & 2/59 (3.39\%) & 3/59 (5.08\%) \\ 
	 Weight decreased & 1/59 (1.69\%) & 3/59 (5.08\%) & 1/59 (1.69\%) & 0/59 (0.0\%) \\ 
	 Heart rate increased & 0/59 (0.0\%) & 4/59 (6.78\%) & 0/59 (0.0\%) & 0/59 (0.0\%) \\ 
	 Fatigue & 2/59 (3.39\%) & 3/59 (5.08\%) & 0/59 (0.0\%) & 0/59 (0.0\%) \\ 
	 Nausea & 6/59 (10.17\%) & 6/59 (10.17\%) & 3/59 (5.08\%) & 1/59 (1.69\%) \\ 
	 Dyspepsia & 2/59 (3.39\%) & 3/59 (5.08\%) & 1/59 (1.69\%) & 0/59 (0.0\%) \\ 
	 Constipation & 1/59 (1.69\%) & 0/59 (0.0\%) & 3/59 (5.08\%) & 1/59 (1.69\%) \\
	
	\bottomrule
	\end{tabular}
\end{table}

\end{document}